\documentclass[10pt, a4paper, conference]{IEEEtran}
\ifCLASSINFOpdf
\else
\fi
\usepackage{times}
\usepackage{epsfig}
\usepackage{graphicx}
\usepackage{amsmath}
\usepackage{amssymb}
\usepackage{multirow}
\usepackage{CJK}

\usepackage{subfig}
\usepackage[colorlinks,linkcolor=red]{hyperref}
\begin{document}
%
\title{Large-scale Isolated Gesture Recognition Using Convolutional Neural Networks}

\author{Pichao Wang$^{\rm 1}$, Wanqing Li$^{\rm 1}$, Song Liu$^{\rm 1}$, Zhimin Gao$^{\rm 1}$, Chang Tang$^{\rm 2}$ and Philip Ogunbona$^{\rm 1}$\\
\\
$^{\rm 1}$Advanced Multimedia Research Lab, University of Wollongong, Australia\\
$^{\rm 2}$School of Information Science and Engineering, Wuhan University of Science and Technology, Wuhan, China\\
\\
{ pw212@uowmail.edu.au, \{wanqing, songl\}@uow.edu.au, zg126@uowmail.edu.au}\\
{ tangchang@wust.edu.cn, philipo@uow.edu.au}
}

\maketitle

\begin{abstract}
This paper proposes three simple, compact yet effective representations of depth sequences, referred to respectively as Dynamic Depth Images (DDI), Dynamic Depth Normal Images (DDNI) and Dynamic Depth Motion Normal Images (DDMNI). These dynamic images are constructed from a sequence of depth maps using bidirectional rank pooling to effectively capture the spatial-temporal information. Such image-based representations enable us to fine-tune the existing ConvNets models trained on image data for classification of depth sequences, without introducing large parameters to learn. Upon the proposed representations, a convolutional Neural networks (ConvNets) based method is developed for gesture recognition and evaluated on the Large-scale Isolated Gesture Recognition at the ChaLearn Looking at People (LAP) challenge 2016. The method achieved 55.57\% classification accuracy and ranked $2^{nd}$ place in this challenge but was very close to the best performance even though we only used depth data.

\end{abstract}


%
\IEEEpeerreviewmaketitle

\begin{figure*}[t]
\begin{center}
{\includegraphics[height = 130mm, width = 180mm]{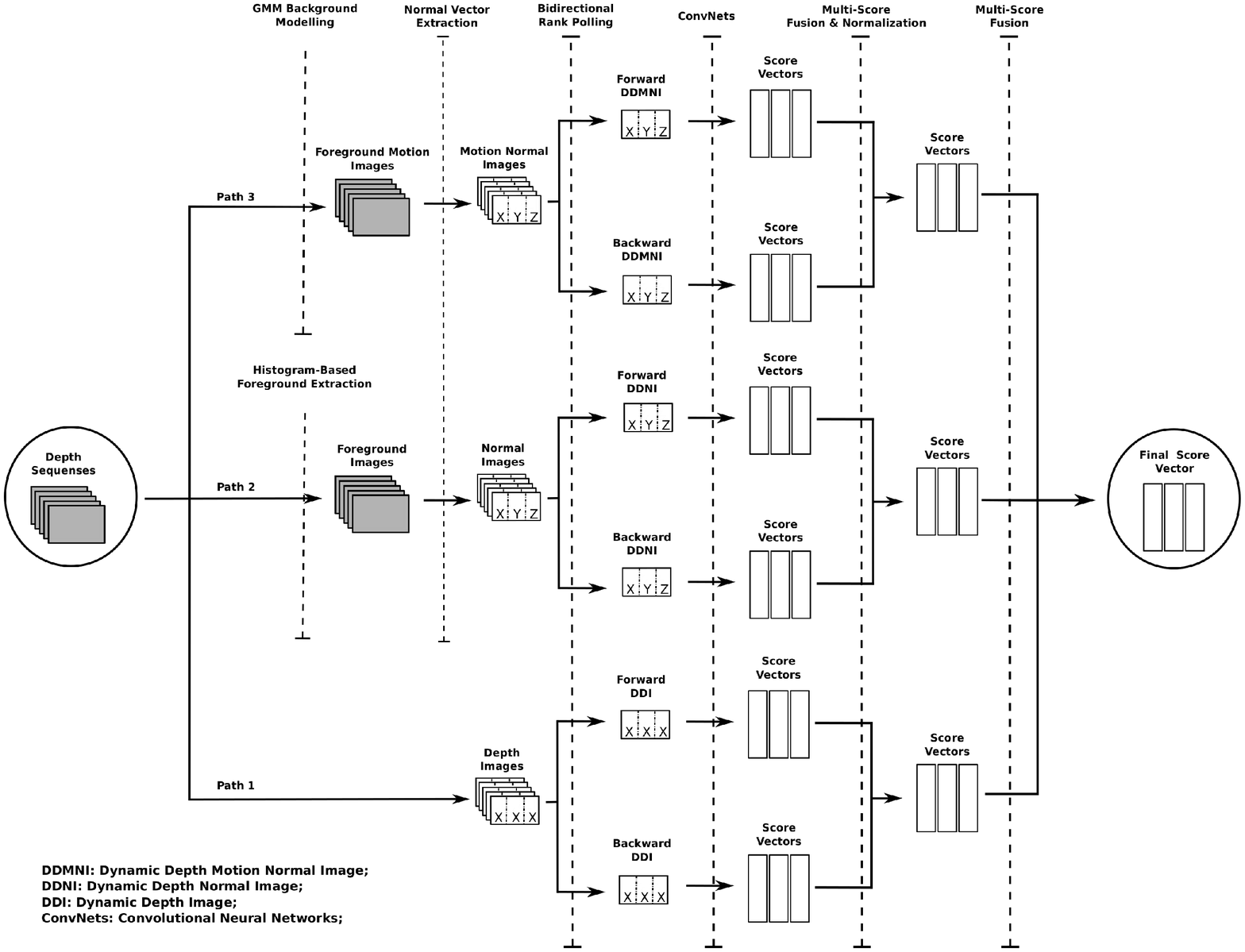}}
\end{center}
\caption{The framework for proposed method.}
\label{fig:framework}
\end{figure*}

\section{Introduction}

Gestures are naturally performed by humans, produced as part of deliberate actions, signs or signals, or subconsciously revealing intentions or attitude~\cite{escalera2016challenges}. While they may involve the motion of all parts of the body, the studies of gestures usually focus on arms and hands which are essential in gesture communication. Recognition of gestures has recently attracted increasing attention due to its indubitable importance in many applications such as Human Computer Interaction (HCI), Human Robot Interaction (HRI) and assistive technologies for the handicapped and the elderly.

Gestures are one type of actions and many action recognition methods can be applied to gesture recognition. Recognition of human actions from depth/skeleton data is one of the most active research topics in multimedia signal processing in recent years due to the advantages of depth information over conventional RGB video, e.g. being insensitive to illumination changes. Since the first work of such a type~\cite{li2010action} reported in 2010, many methods~\cite{wang2012mining,Yang2012a,Oreifej2013,Gowayyed2013_HOD,yangsuper,rahmani2014hopc,
pichao2014,lurange,escalante2015principal,Vemulapallia2016,zhang2016rgb} have been proposed based on specifical hand-crafted feature descriptors extracted from depth/skeleton. With the recent development of deep learning, a few methods have been developed based on Convolutional Neural Networks (ConvNets)~\cite{pichao2015,pichaoTHMS,pichao2016,pichaoicprwa,pichaocsvt2016} and Recurrent Neural Networks (RNNs)~\cite{du2015hierarchical,veeriah2015differential,zhu2015co,shahroudy2016ntu}.  However, it remains unclear how video could be effectively represented and fed to deep neural networks for classification. For example, one can conventionally consider a video as a sequence of still images with some form of temporal smoothness, or as a subspace of images or image features, or as the output of a neural network encoder. Which one among these and other possibilities would result in the best representation in the context of gesture recognition is not well understood. 


Inspired by the recent work in~\cite{pichao2015,pichaoTHMS,pichao2016,bilen2016dynamic}, this paper proposes for gesture recognition three simple, compact and effective representations of depth sequences which effectively decribe a short depth sequence with images. Such representations make it possible to use a standard ConvNet architecture to learn suitable ``dynamic" features from the sequences by utilizing the ConvNet models trained from image data. Consequently, it avoids training millions of parameters from scratch and is especially valuable in the cases that lack sufficient annotated training video data. For instance, the large-scale isolated gesture recognition challenge~\cite{wanchalearn} has on average only 144 video clips per class compared to 1200 images per class in ImageNet.


The proposed three representations are Dynamic Depth Image (DDI), Dynamic Depth Normal Image (DDNI) and Dynamic Depth Motion Normal Image (DDMNI). They are all constructed from a sequence of depth maps based on bidirectional rank pooling to encode the spatial (i.e. posture) and temporal (i.e. motion) information at different levels and are complementary to each other. Experimental results have shown that the three representations can improve the recognition accuracy substantially. 

The rest of this paper is organized as follows. Section II briefly reviews the related works on gesture/action recognition based on depth and deep learning. Details of the proposed method are described in Section III. Experimental results are presented in Section IV. Section V concludes the paper.

\section{Related Work}
\subsection{Depth Based Action Recognition}
With Microsoft Kinect Sensors researchers have developed methods for depth map-based action recognition. Li et al. \cite{li2010action} sampled points 
from a depth map to obtain a bag of 3D points to encode spatial information and 
employ an expandable graphical model to encode temporal information 
\cite{li2008}. Yang et al. \cite{Yang2012a} stacked differences between projected depth maps as 
a depth motion map (DMM) and then used HOG to extract relevant features from the 
DMM. This method transforms the problem of action recognition from 
spatio-temporal space to spatial space.  In \cite{Oreifej2013}, a feature called Histogram of Oriented 4D 
Normals (HON4D) was proposed; surface normal is extended to 4D space and 
quantized by regular polychorons. Following this method, Yang and Tian 
\cite{yangsuper} cluster hypersurface normals and form the polynormal which can 
be used to jointly capture the local motion and geometry information. Super 
Normal Vector (SNV) is generated by aggregating the low-level polynormals. In 
\cite{lurange}, a fast binary range-sample feature was proposed based on a test 
statistic by carefully designing the sampling scheme to exclude most pixels that 
fall into the background and to incorporate spatio-temporal cues.

\subsection{Deep Leaning Based Recognition}
Exiting deep learning approach can be generally divided into four categories based on how the video is represented and fed to a deep neural network. The first category views a video either as a set of still images~\cite{yue2015beyond} or as a short and smooth transition between similar frames~\cite{simonyan2014two}, and each color channel of the images is fed to one channel of a ConvNet. Although obviously suboptimal, considering the video as a bag of static frames performs reasonably well. The second category is to represent a video as a volume and extends ConvNets to a third, temporal dimension~\cite{ji20133d,tran2015learning} replacing 2D filters with 3D ones. So far, this approach has produced little benefits, probably due to the lack of annotated training data. The third category is to treat a video as a sequence of images and feed the sequence to a RNN~\cite{donahue2015long,du2015hierarchical,veeriah2015differential,zhu2015co,shahroudy2016ntu}. A RNN is typically considered as memory cells, which are sensitive to both short as well as long term patterns. It parses the video frames sequentially and encode the frame-level information in their memory. However, using RNNs did not give an improvement over temporal pooling of convolutional features~\cite{yue2015beyond} or over hand-crafted features. The last category is to represent a video in one or multiple compact images and adopt available trained ConvNet architectures for fine-tuning~\cite{pichao2015,pichaoTHMS,pichao2016,bilen2016dynamic}. This category has achieved state-of-the-art results of action recognition on many RGB and depth/skeleton datasets. The proposed method in this paper falls into the last category.

\section{Proposed Method}

The proposed method consists of three stages: construction of the three sets of dynamic images, ConvNets training and score fusion for classification, as illustrated in Fig.~\ref{fig:framework}. Details are presented in the rest of this section. 

\subsection{Construction of Dynamic Images}

The three sets of dynamic images, Dynamic Depth Images (DDIs), Dynamic Depth Normal Images (DDNIs) and Dynamic Depth Motion Normal Images (DDMNIs) are constructed from a sequence of depth maps through rank pooling~\cite{bilen2016dynamic}. They aim to capture both posture and motion information for gesture recognition.

\subsubsection{Rank Pooling}

Let $I_{1},...,I_{T}$ denote the frames in a sequence of depth maps, and $\varphi(I_{t}) \in \mathbb{R}^{d}$ be a representation or feature vector extracted from each individual frame $I_{t}$. Let $V_{t} = \dfrac{1}{t}\sum_{\tau=1}^{t}\varphi(I_{t})$ be time average of these features up to time $t$. The ranking function associates to each time $t$ a score $S(t|\textbf{d}) = <\textbf{d}, V_{t}>$, where $\textbf{d} \in \mathbb{R}^{d}$ is a vector of parameters. The function parameters $\textbf{d}$ are learned so that the scores reflect the rank of the frames in the video. In general, later times are associated with larger scores, $ i.e. ~q > t \Rightarrow S(q|\textbf{d}) > S(t|\textbf{d})$. Learning $\textbf{d}$ is formulated as a convex optimization problem using RankSVM~\cite{smola2004tutorial}:

\begin{equation}
\begin{aligned}
\textbf{d}^{*} &=\rho(I_{1},...,I_{T};\varphi) = \mathop{\arg\min}_{\textbf{d}} E(\textbf{d}),\\
  E(\textbf{d}) &= \dfrac{\lambda}{2}\parallel \textbf{d} \parallel^{2} +\\
  & \dfrac{2}{T(T-1)}\times\sum\limits_{q>t}max\{{0,1-S(q|\textbf{d}) + S(t|\textbf{d})}\}.
\end{aligned}
\end{equation}

The first term in this objective function is the usual quadratic regular term used in SVMs. The second term is a hinge-loss soft-counting how many pairs $q > t$ are incorrectly ranked by the scoring function. Note in particular that a pair is considered correctly ranked only if scores are separated by at least a unit margin, $i.e.~ S(q|\textbf{d}) > S(t|\textbf{d}) + 1$. 

Optimizing the above equation defines a function $\rho(I_{1},...,I_{T};\varphi)$ that maps a sequence of $T$ depth video frames to a single vector $d^{*}$. Since this vector contains enough information to rank all the frames in the video, it aggregates information from all of them and can be used as a video descriptor. This process is called rank pooling. 

\subsubsection{Construction of DDI}
Given a sequence of depth maps, the ranking pooling method~\cite{bilen2016dynamic} described above is employed to generate a dynamic depth image (DDI).  The DDI is fed to the three channel of a ConvNet. Different from~\cite{bilen2016dynamic} the rank pooling is applied in a bidiretional way to convert one depth map sequence into two DDIs. As shown in Fig.~\ref{fig:DIs}, DDIs effectively capture the posture information, similar to key poses.

\subsubsection{Construction of DDNI}

In order to simultaneously exploit the posture and motion information in depth sequences, it is proposed to extract normals from depth maps and construct the so called DDNIs (dynamic depth normal images). For each depth map, the surface normal $(n_x,n_y,n_z)$ at each location is calculated. Thus, three channels $(N_x,N_y,N_z)$, referred to as a Depth Normal Image (DNI), are generated from the calculated normals, where $(N_x,N_y,N_z)$ represents normal images for the three components $(n_x,n_y,n_z)$ respectively. The sequence of DNIs goes through bidirectional rank pooling to generate two DDNIs, one  being from forward ranking pooling and the other from backward rank pooling.

To minimise the interference of the background, it is assumed that the background in the histogram of depth maps occupies the last peak representing far distances. Specifically, pixels whose depth values are greater than a threshold defined by the last peak of the depth histogram minus a fixed tolerance (0.1 was set in our experiments) are considered as background and removed from the calculation of DDNIs by re-setting their depth values to zero. Through this simple process, most of the background can be removed and has much contribution to the DDNIs.  Samples of DDNIs can be seen in Fig.~\ref{fig:DIs}.

\subsubsection{Construction of DDMNI}

\begin{figure}[t]
\begin{center}
{\includegraphics[height = 120mm, width = 85mm]{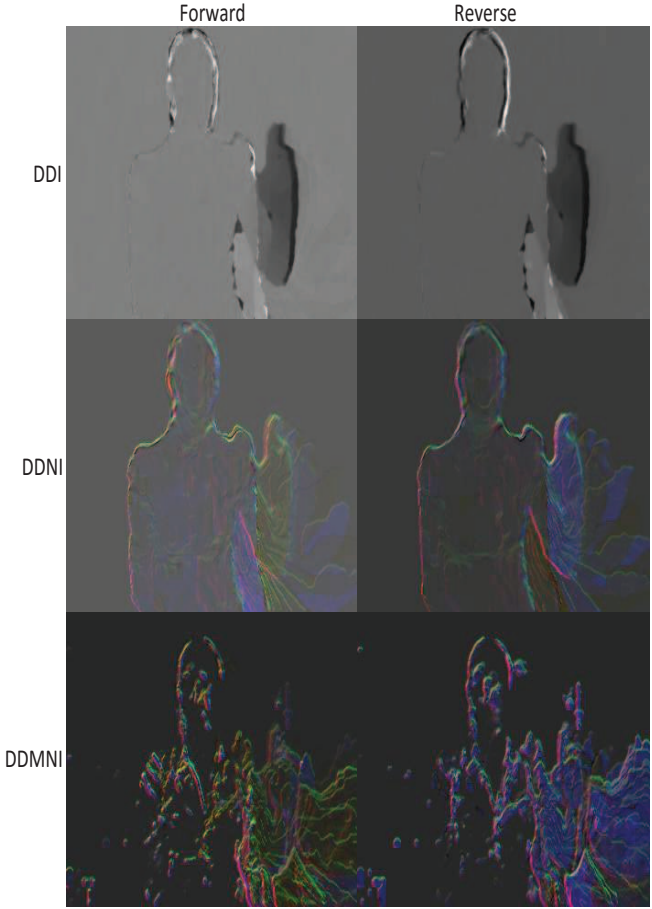}}
\end{center}
\caption{Samples of generated forward and backward DDIs, DDNIs and DDMNIs for gesture Mudra1/Ardhapataka.}
\label{fig:DIs}
\end{figure}

The purpose of construction of a DDMNI is to further exploit the motion in depth maps. Gaussian Mixture Models (GMM) is applied to depth sequences to detect moving foreground. The same process as the construction of a DDNI ( but without using histogram-based foreground extraction) is employed to the moving foreground. This process generates two DDMNIs, which specifically capture the motion information as illustrated in Fig.~\ref{fig:DIs}.

\subsection{Network Training}
After the construction of DDIs, DDNIs and DDMNIs, there are six dynamic images, as illustrated in Fig.~\ref{fig:DIs}, for each depth map sequence. Six ConvNets were trained on the six channels individually. Different layer configurations were used for the validation and testing sets provided by the Challenge. For validation, the layer configuration of six ConvNets follows the one in~\cite{krizhevsky2012imagenet}.  For testing, VGG-16~\cite{simonyan2014very} was adopted for fine-tuning. The implementation is derived from the publicly available Caffe toolbox~\cite{jia2014caffe} based on three {NVIDIA Tesla K40 GPU} cards for both validation and testing.

The training procedure for validation is similar to the one in~\cite{krizhevsky2012imagenet}. The network weights were learned using the mini-batch stochastic gradient descent with the momentum being set to 0.9 and weight decay being set to 0.0005. All hidden weight layers use the rectification (RELU) activation function. At each iteration, a mini-batch of 256 samples is constructed by sampling 256 shuffled training samples. All the images are resized to 256 $\times$ 256. The learning rate was set to $10^{-3}$ for fine-tuning with pre-trained models on ILSVRC-2012, and then it is decreased according to a fixed schedule, which is kept the same for all training sets. For each ConvNet the training undergoes 20K iterations and the learning rate decreases every 5K iterations. For all experiments, the dropout regularisation ratio was set to 0.5 in order to reduce complex co-adaptations of neurons in the nets.

For testing, the training procedure is similar to the one in~\cite{simonyan2014very}. The network weights were learned using the mini-batch stochastic gradient descent with the momentum being set to 0.9 and weight decay being set to 0.0005. All hidden weight layers use the rectification (RELU) activation function. At each iteration, a mini-batch of 32 samples was constructed by sampling 256 shuffled training samples. All the images are resized to 224 $\times$ 224. The learning rate was set to $10^{-3}$ for fine-tuning with pre-trained models on ILSVRC-2012, and then it is decreased according to a fixed schedule, which is kept the same for all training sets. For each ConvNet the training undergoes 50K iterations and the learning rate decreases every 20K iterations. For all experiments, the dropout regularisation ratio was set to 0.9 in order to reduce complex co-adaptations of neurons in the nets.

\subsection{Score Fusion for Classification}

Given a testing depth video sequence (sample), three pairs of dynamic images (DDIs, DDNIs, DDMNIs) are generated and fed into six different trained ConvNets. For each image pair, multiply-score fusion was used. The score vectors outputted by the two pair ConvNets are multiplied in an element-wise  way and then the resultant score vectors are normalized using $L_{1}$ norm. The three normalized score vectors are then multiplied in an element-wise fashion and the max score in the resultant vector is assigned as the probability of the test sequence being the recognized class. The index of this max score corresponds to the recognized class label.

\begin{table*}[!ht]
\centering
\caption{Information of the ChaLearn LAP IsoGD Dataset. \label{table1}}
\begin{tabular}{|c|c|c|c|c|c|c|}
\hline
Sets &\# of labels &\# of gestures & \# of RGB videos & \# of depth videos & \# of subjects & label provided \\
\hline
Training & 249 & 35878 & 35878 & 35878 & 17 & Yes \\
\hline
Validation & 249 & 5784 & 5784 & 5784 & 2 & No \\
\hline
Testing & 249 &  6271 & 6271 & 6271 & 2 & No \\
\hline
All & 249 & 47933 & 47933 & 47933 & 21 & - \\
\hline
\end{tabular}
\end{table*}

\section{Experiments}
In this section, the Large-scale Isolated Gesture Recognition Dataset at the ChaLearn LAP challenge 2016 (ChaLearn LAP IsoGD Dataset)~\cite{ICPRW2016} and the evaluation protocol are described. The experimental results of the proposed method on this dataset are presented.

\subsection{Dataset}
\begin{figure*}[t]
\begin{center}
{\includegraphics[height = 70mm, width = 180mm]{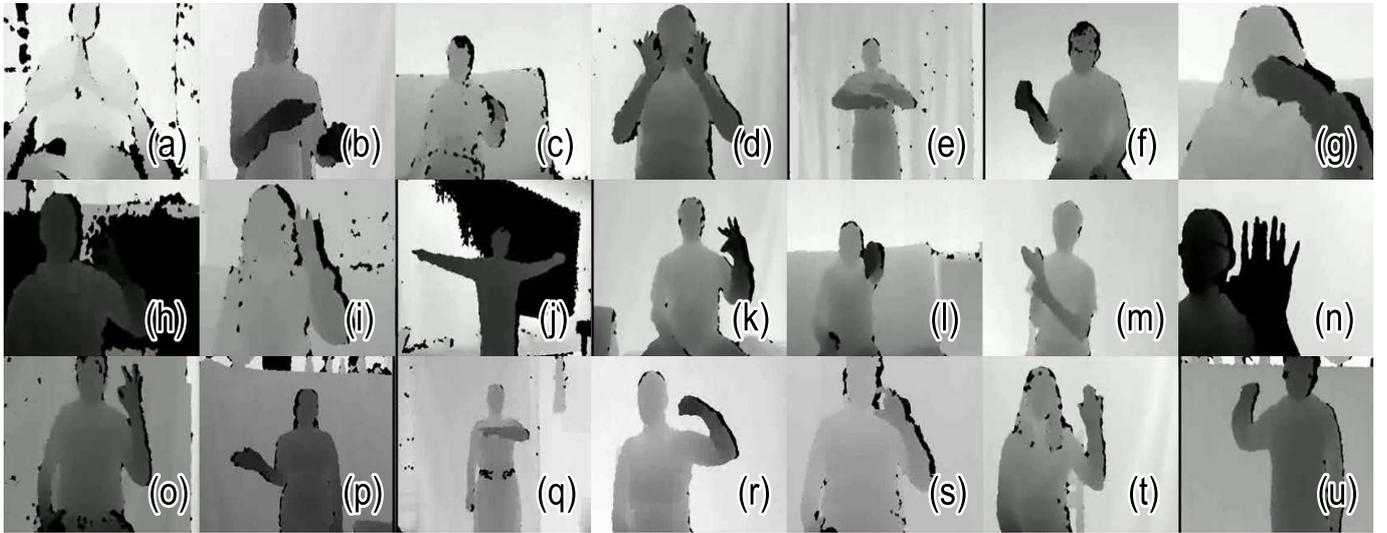}}
\end{center}
\caption{The samples of 21 out of 249 gestures. From top left to bottom right, they are: \\(a) ItalianGestures/Madonna; (b) GestunoTopography/92\_harbour\_port; (c) TaxiSouthAfrica/TaxiHandSigns2;\\ (d) GestunoSmallAnimals/129\_cat\_chat;(e) RefereeWrestlingSignals2/Reversal; (f) DivingSignals3/NotUnderstood;\\(g) SurgeonSignals/ArmyNavyRetractor; (h) GangHandSignals1/EastSide; (i) SwatHandSignals1/DogNeeded;\\(j) HelicopterSignals/MoveLeft; (k) GangHandSignals2/Killas; (l) TaxiSouthAfrica/TaxiHandSigns6;\\ (m) DivingSignals4/HowMuchAir; (n) ChineseNumbers/wu,TaxiSouthAfrica/TaxiHandSigns7;\\ (o) Mudra2/Vitarka,DivingSignals4/OK,GangHandSignals2/OK; (p) DivingSignals1/Around;\\(q) CanadaAviationGroundCirculation1/DirigezVousVers; (r) MusicNotes/do; (s) GangHandSignals1/Crip;\\ (t) SwatHandSignals1/Stop; (u) RefereeWrestlingSignals2/Stalling,SwatHandSignals1/Breacher.}
\label{fig:samples}
\end{figure*}

The ChaLearn LAP IsoGD Dataset is derived from the ChaLearn Gesture Dataset (CGD)~\cite{guyon2014chalearn}. It includes 47933 RGB-D depth sequences, each RGB-D video representing one gesture instance. There are 249 gestures performed by 21 different individuals.  The detailed information of this dataset are shown in Table~\ref{table1}.  In this paper, only depth maps were used to evaluate the performance of the proposed method. Some samples of depth sequences are shown in Fig.~\ref{fig:samples}.

\subsection{Evaluation Protocal}

The dataset is divided into training, validation and test sets. All three sets consist of samples of different subjects so ensure that the gestures of one subject in validation and test sets will not appear in the training set. 

For the isolated gesture recognition challenge, recognition rate $r$ is used as the evaluation criteria. The recognition rate is calculated as:
\begin{equation}
r = \dfrac{1}{n}\delta(p_{l}(i),t_{l}(i))
\end{equation}
where $n$ is the number of samples; $p_{l}$ is the predicted label; $t_{l}$ is the ground truth; $\delta(j_{1},j_{2}) = 1$, if $j_{1} = j_{2}$, otherwise $\delta(j_{1},j_{2}) = 0$.

\subsection{Experimental Results}

The results obtained by the proposed method on the validation and test sets are listed and compared with the baseline methods~\cite{pami16Jun} (MFSK and MFSK+DeepID) in Table~\ref{table2}. The codes and models can be downloaded at the author's homepage:\url{https://sites.google.com/site/pichaossites/}\href{https://sites.google.com/site/pichaossites/}.

\begin{table}[!ht]
\centering
\caption{Comparative accuracy of proposed method and baseline 
methods on the ChaLearn LAP IsoGD Dataset. \label{table2}}
\begin{tabular}{|c|c|c|}
\hline
Method & Set & Recognition rate $r$\\
\hline
MFSK & Validation & 18.65\%\\
\hline
MFSK+DeepID & Validation & 18.23\%\\
\hline
Proposed Method & Validation & \textbf{39.23\%}\\
\hline
MFSK & Testing & 24.19\% \\
\hline
MFSK+DeepID & Testing & 23.67\%\\
\hline
Proposed & Testing &\textbf{55.57\%}\\
\hline
\end{tabular}
\end{table}

The results showed that the proposed method 
significantly outperformed the baseline methods, even though only single 
modality, i.e. depth data, was used while the baseline method used both RGB and 
depth videos.

The challenge results are summarized in Table~\ref{table3}. We can see that our method is among the top performers and our recognition rate is very close to the best performance of this challenge (55.5733\% vs. 56.8968\%), even though we only used depth data for proposed method while the winner~\cite{yunanli} adopted both depth and RGB modalities.

\begin{table}[!ht]
\centering
\caption{Comparsion the performance of our submission with those of other teams. Our team secures the second place in the ICPR ChaLearn LAP challenge 2016. \label{table3}}
\begin{tabular}{|c|c|c|}
\hline
Rank & Team & Recognition rate $r$\\
\hline
1 & FLiXT~\cite{yunanli} & 56.8968\%\\
\hline
2 & \textbf{AMRL (ours)} & 55.5733\%\\
\hline
3 & XDETVP-TRIMPS~\cite{guangming} & 50.9329\%\\
\hline
4 & ICT\_NHCI & 46.8027\%\\
\hline
5 & XJTUfx & 43.9164\%\\
\hline
6 & TARDIS & 40.1531\%\\
\hline
7 & NTUST & 20.3317\%\\
\hline
\end{tabular}
\end{table}

\section{Conclusions}

This paper presented three simple, compact yet effective representations of depth sequences for gesture recognition using convolutional Neural networks. They are all based on bidirectional rank pooling method converting the depth sequences into images. Such representations enables the use of existing ConvNets models directly on video data with fine-tuning without introducing large parameters to learn. The three representations represent the posture and motion in different levels and they are complementary to each other and improve the recognition accuracy largely. Experimental results on ChaLearn LAP IsoGD Dataset verified the effectiveness of the proposed method.

\section*{Acknowledgment}
The authors would like to thank NVIDIA Corporation for the donation of a Tesla K40 GPU card used in this challenge.


\end{document}